\documentclass{ecai}
\usepackage{graphicx}
\usepackage{latexsym}

\usepackage{hyperref}
\usepackage{url}
\usepackage{graphicx}
\usepackage{algorithm}
\usepackage{algpseudocode}
\usepackage{multirow}
\usepackage{xcolor}
\usepackage{booktabs}
\usepackage{caption}
\usepackage{wrapfig}
\usepackage{array}
\usepackage[square,sort,comma,numbers]{natbib}
\usepackage{stfloats}
\usepackage{mathtools}
\usepackage{amsfonts}

\renewcommand*\thefootnote{\arabic{footnote}}
\begin{document}

\begin{frontmatter}

\title{Preserving Semantics in Textual \\ Adversarial Attacks}

\author[A, B]{\fnms{David}~\snm{Herel} 
\orcid{0009-0000-1861-3778} 
\thanks{Corresponding Author. Email: hereldav@fel.cvut.cz}}
\author[B]{\fnms{Hugo}~\snm{Cisneros} 
\orcid{0000-0003-3439-4565}
}
\author[B]{\fnms{Tomas}~\snm{Mikolov}
\orcid{0000-0002-6938-5426}
}

\address[A]{Faculty of Electrical Engineering, Czech Technical University in Prague }
\address[B]{Czech Institute of Informatics, Robotics and Cybernetics, Czech Technical University in Prague}

\begin{abstract}
The growth of hateful online content, or hate speech, has been associated with a global increase in violent crimes against minorities \cite{hate_speech_definition}. Harmful online content can be produced easily, automatically and anonymously. Even though, some form of auto-detection is already achieved through text classifiers in NLP, they can be fooled by adversarial attacks. To strengthen existing systems and stay ahead of attackers, we need better adversarial attacks. In this paper, we show that up to 70\% of adversarial examples generated by adversarial attacks should be discarded because they do not preserve semantics. We address this core weakness and propose a new, fully supervised sentence embedding technique called Semantics-Preserving-Encoder (SPE). Our method outperforms existing sentence encoders used in adversarial attacks by achieving 1.2$\times$ $\sim$ 5.1$\times$ better real attack success rate. We release our code as a plugin that can be used in any existing adversarial attack to improve its quality and speed up its execution\footnotemark.
\vspace{-8pt}
\end{abstract}

\end{frontmatter}

\section{Introduction}
\label{sec:intro}

\begin{figure}[ht!]
\includegraphics[width=\linewidth]{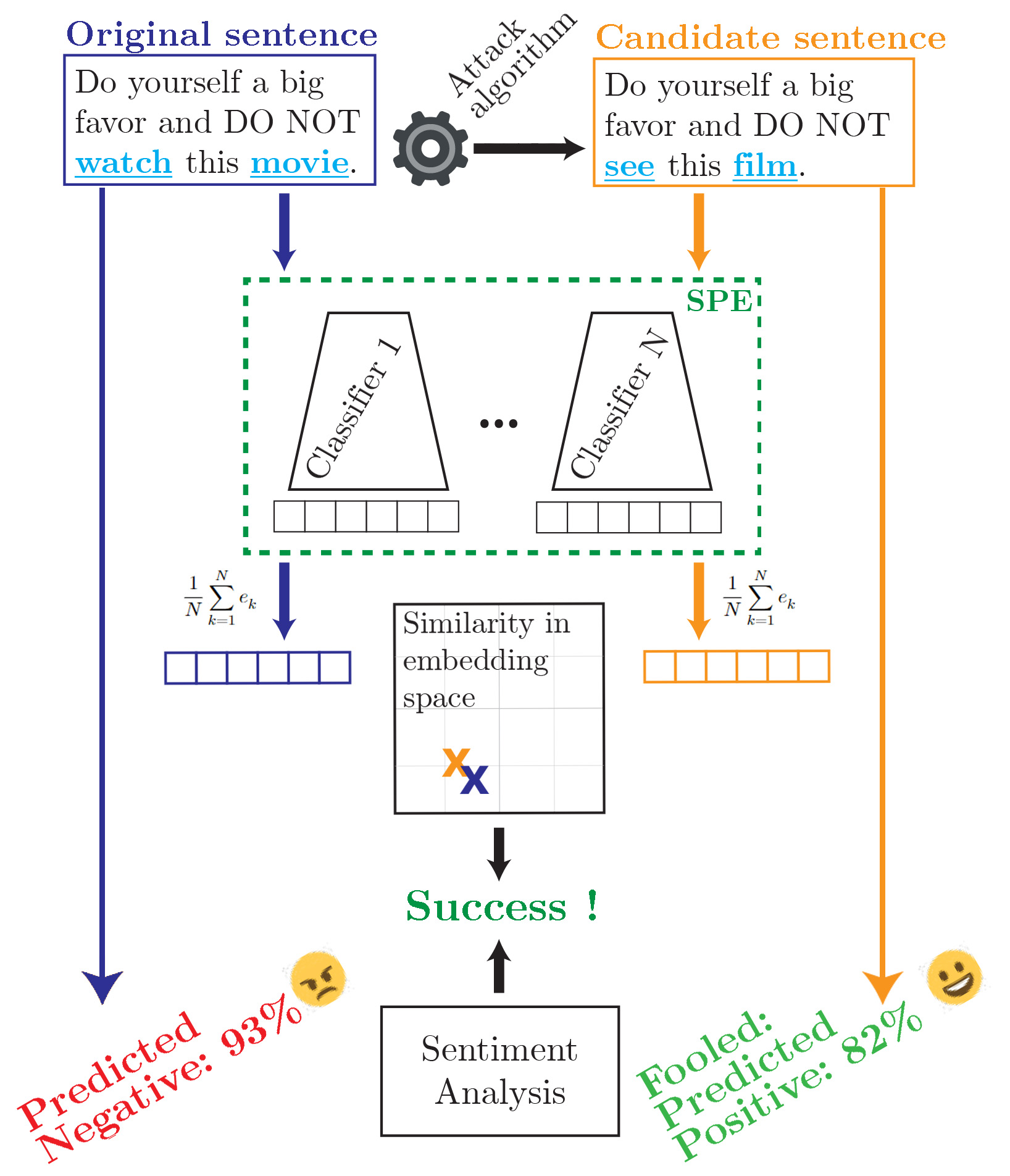}
\caption{A high-level overview of textual adversarial attack and our proposed Semantic-Preserving-Encoder (SPE). Our method takes advantage of supervised learning to tackle the problem of preserving semantics.}
\label{fig:visual_abs}
\end{figure}

With the massive growth of online content, the need to detect harmful content, such as hate speech, offensive language, or fake news, is increasingly important. Current defense system relies on deep learning models in NLP. However, many researchers have highlighted that language models are not as robust as previously thought \cite{jinBERTReallyRobust2020,cv_adv,zhao2018generating,intrig_proper,at_phys,at_scale} and that they can be fooled quite easily with so-called adversarial examples. Adversarial examples introduce a small perturbation to the input data that is 'imperceptible' to the human eye and fools the system into misclassifying the perturbed data. For example, in the domain of offensive language detection, we can have an offensive text on input and modify it in such a way that the meaning is preserved, but the modified text will fool the system to classify it as non-offensive \cite{jinBERTReallyRobust2020}. A similar scenario is illustrated in Figure \ref{fig:visual_abs} in the domain of sentiment analysis of movie reviews. \par
It is important to mention that the imperceptibility of adversarial attacks in discrete domains such as text is much more difficult to define than in a continuous space, since in discrete domains an indistinguishable modification simply cannot exist.
According to \cite{jinBERTReallyRobust2020,adv_quality_morris} we can identify three main requirements for an adversarial attack on text to be successful:
1. Human prediction consistency: The prediction made by humans should remain unchanged; 
2. Semantic similarity: The designed example should have the same meaning as the source as judged by humans; 
3. Language fluency: the generated examples should look natural and grammatical.
\par
\footnotetext[1]{The code, datasets and test examples are available at \url{https://github.com/DavidHerel/semantics-preserving-encoder}. \label{fnlabel}}

Although most adversarial attacks claim to meet these constraints \cite{bl_box_at, li-etal-2021-contextualized, garg-ramakrishnan-2020-bae}, careful scrutiny shows otherwise. We observe that many adversarial examples do not preserve the meaning of the text. This is also supported by Morris et al. \cite{adv_quality_morris} whose findings are similar. To overcome this problem, Morris et al. \cite{adv_quality_morris} suggested increasing cosine thresholds and introducing mechanisms such as grammar checks to improve the quality of adversarial examples. However, it is at the cost of the attack success rate, which decreased rapidly by more than 70\%. We suggest a different solution that promises to avoid this decline in the attack success rate. \par
It appears that the problem lies in the similarity metric itself, whose function is to measure the difference between the original and perturbed sentences. These metrics mostly use encoders that are trained with limited supervision. This makes them more susceptible to problems with antonym recognition, as illustrated in the third column of Table \ref{table:use_spe}.

We propose a new sentence encoder for similarity metrics in textual adversarial attacks called Semantics-Preserving-Encoder (SPE). Our method takes advantage of supervised learning on annotated datasets. Thus, it is more robust towards the antonym recognition problems that are observed in adversarial examples. Experimental results show that our proposed method outperforms existing sentence encoders used in adversarial attacks by achieving 1.2$\times$ $\sim$ 5.1$\times$ better real attack success rate, while also being significantly faster. Furthermore, our solution -- SPE is publicly available on GitHub\footnote[2]{The code, datasets and test examples are available at \url{https://github.com/DavidHerel/semantics-preserving-encoder}.} and can be integrated as a component into any existing adversarial attack granting much faster execution. 

The contributions of this work can be summarized as follows:
\begin{enumerate}
  \item We propose a new sentence-encoder technique, SPE, for textual adversarial attacks. Our method outperforms existing sentence encoders used in adversarial attacks, achieving 1.2$\times$ $\sim$ 5.1$\times$ better real attack success rate.

  \item We propose a new metric for evaluating the quality of the adversarial examples -- rASR, which reflects the real performance of adversarial attacks.

  \item We evaluate some of the most common sentence encoders used in adversarial attacks both manually and automatically on relevant datasets such as hate-speech and offensive language detection \cite{barbieriTweetEvalUnifiedBenchmark2020}, in addition to other popular classification tasks such as Yelp Reviews \cite{zhangCharacterlevelConvolutionalNetworks2015} and Rotten Tomatoes \cite{pangSeeingStarsExploiting2005}.
  
  \item We release our work as open source, including the code, human evaluations, datasets, and test samples for the purpose of reproducibility and future benchmarking.

\end{enumerate}

\section{Related work}
\textbf{Textual adversarial attacks.} There were many attempts to create textual adversarial attacks that preserve semantics and are grammatically correct. Ultimately, we can distinguish between three different ways in which other researchers approach this. The first approach aims to modify the whole sentence using various sophisticated phrase perturbations, such as paraphrasing \cite{gan-ng-2019-improving, wang-etal-2020-t3}. However, these modifications often have problems with semantic preservation \cite{Zhang2019GeneratingTA}. \par
The second approach focuses on character-level modification, such as misspellings or typos, which has proven to be more successful in terms of semantics preservation \cite{he-etal-2021-model, Li2019TextBuggerGA, ebrahimi-etal-2018-hotflip}. However, research shows that these types of attacks can be mitigated quite easily with tools such as grammar checks \cite{pruthi-etal-2019-combating, jones-etal-2020-robust}. \par
Lastly, the word-level attack technique focuses on the substitution or modification of a single word (or combination of multiple words) in the text \cite{rew_1, rew_2, rew_3, clare}. This type of attack aims to preserve the constraints defined by \cite{morris-etal-2020-reevaluating}, often using methods such as synonym substitution to improve semantic preservation.\par
\textbf{Similarity metrics.} The majority of word-level adversarial attacks enforce semantics similarity by using Universal-Sentence-Encoder (USE) \cite{USE} or BERTScore \cite{zhangBERTScoreEvaluatingText2020} encoders, both of which are trained mainly on unsupervised tasks.
USE is trained on a task such as Skip-Thought \cite{skip-though}, a conversational input-response task \cite{Henderson} and a supervised classification task performed on the SNLI dataset \cite{snli}. BERTScore is based on a pre-trained BERT language model \cite{bert}, which was also trained unsupervisedly on the Next Sentence Prediction and Masked LM tasks. \par
Even though these sentence encoders have been thoroughly studied on various general tasks, only a few previous works acknowledge their flaws when used in adversarial attacks \cite{morris-etal-2020-reevaluating, david2022adversarial}. In most cases, these encoders struggle to recognize changes in the text's meaning and semantics. To overcome this, Morris et al. \cite{morris-etal-2020-reevaluating} increased the cosine threshold, resulting in an improved quality of adversarial examples, but over 70\% decline in the attack success rate.\par
Looking at the big picture of the observed problems, we have discovered a potential link to the encoder training process. We show that the unsupervised training predetermines these encoders to have problems with antonym recognition, which leads us to introduce our own Semantics-Preserving-Encoder.\par

\section{Method}
Problems of textual adversarial attacks are formulated in the following section. Next, SPE is formally introduced together with its classifiers and other properties. \par

\label{sec:method}
\subsection{Problem formulation}

Most similarity metrics in adversarial attacks rely on sentence encoders such as USE \cite{USE} or BERTScore \cite{zhangBERTScoreEvaluatingText2020}. These encoders use multi-task learning with a high emphasis on unsupervised tasks. The BERT language model \cite{bert} uses Masked-Language Modeling. USE is trained on Skip-Thought \cite{skip-though} like task, where the goal is to predict the middle sentence based on the given context. Both of these training methods could be seen as a variation of a Skip-Gram/CBOW model \cite{skip-gram}, where the goal is to predict the context of a given target word, and vice versa for CBOW. However, this training method forces synonyms to be mapped into a similar vector space as antonyms, since they appear in the same context. Therefore, two contradictory sentences in their vector representation could be very close to each other in the vector space despite their opposite meaning. That is if the unsupervised training with Skip-Gram or CBOW like tasks is used. This is the case for both BERT and USE which is shown in Table \ref{table:use_spe}. \par 

\subsection{Semantics-Preserving-Encoder}
The core idea of our encoder lies in supervised training. We took advantage of existing prelabeled datasets and utilized them in the training data to tackle the problem with opposite words appearing in the same context. As a result, the words that are the most discriminative for the given label will be close to each other in the vector space. Thus, sentences like \textit{'This movie is so good'} and \textit{'This movie is so bad'} should never be close to each other in the vector space, because their semantics label will be exactly opposite. 

We have combined $N$ classifiers trained on different annotated datasets, allowing us to have a diverse set of different sentence vectors. The sentence vectors will differ because each classifier produces its vector according to the task on which it was trained. Therefore, the diversity of classifiers implies a diverse set of sentence vectors. Moreover, by combining several sentence embeddings from different classifiers, we can create a robust classifier that can produce a high-quality embedding for a broad range of topics. 

From a sentence $S$, an attack will generate a candidate adversarial example $S^*$. We denote the $N$ supervised classifiers by $C_1$, $C_2$, \ldots, $C_N$. For simplicity, we consider these classifiers to be functions whose outputs are the sentence embeddings extracted from the classifier when applied to a sentence.
Formally, for all $k$, $C_k(S) = \boldsymbol{e}_k \in \mathbb{R}^p$ where $p$ is the embedding dimension. The complete embedding of a sentence is obtained by averaging the output of the classifiers into a single embedding vector. Average helps to reduce noise and improve the overall accuracy of a model by smoothing out the influence of any individual data points that may be outliers or contain errors.
\[
\boldsymbol{e}^{(S)} = \frac{1}{N} \sum_{k = 1}^{N} C_k(s) = \frac{1}{N} \sum_{k = 1}^{N} \boldsymbol{e}_k^{(S)}. 
\]
The similarity between the original sentence $S$ and the attacked sentence $S^*$ is computed with the cosine similarity between their embeddings as follows
\[
\text{Sim}(S, S^*) = \frac{\boldsymbol{e}^{(S)} \cdot \boldsymbol{e}^{(S^*)}}{\|\boldsymbol{e}^{(S)}\|\|\boldsymbol{e}^{(S^*)}\|}
,\]
where $\cdot$ represents the dot product between vectors in $\mathbb{R}^p$ and $\|\cdot\|$ is the $L_2$ norm in $\mathbb{R}^p$.
For a threshold $\epsilon$, $S$ and $S^*$ will be considered to have the same meaning if $\text{Sim}(S, S^*) > \epsilon$.

The classification model that we used is fastText \cite{fasttext}. However, it is important to note that any other classification model can be used instead. We decided to use fastText due to its many advantages. Firstly, fastText classifiers allow us to create sentence vectors rather quickly with a reasonable performance for the given task. Secondly, we can reduce the dimensionality of the vector space. This way we can put more information into fewer dimensions, which results in a more efficient space storage. 

\begin{figure}[htbp]
\centering
\includegraphics[width=\linewidth]{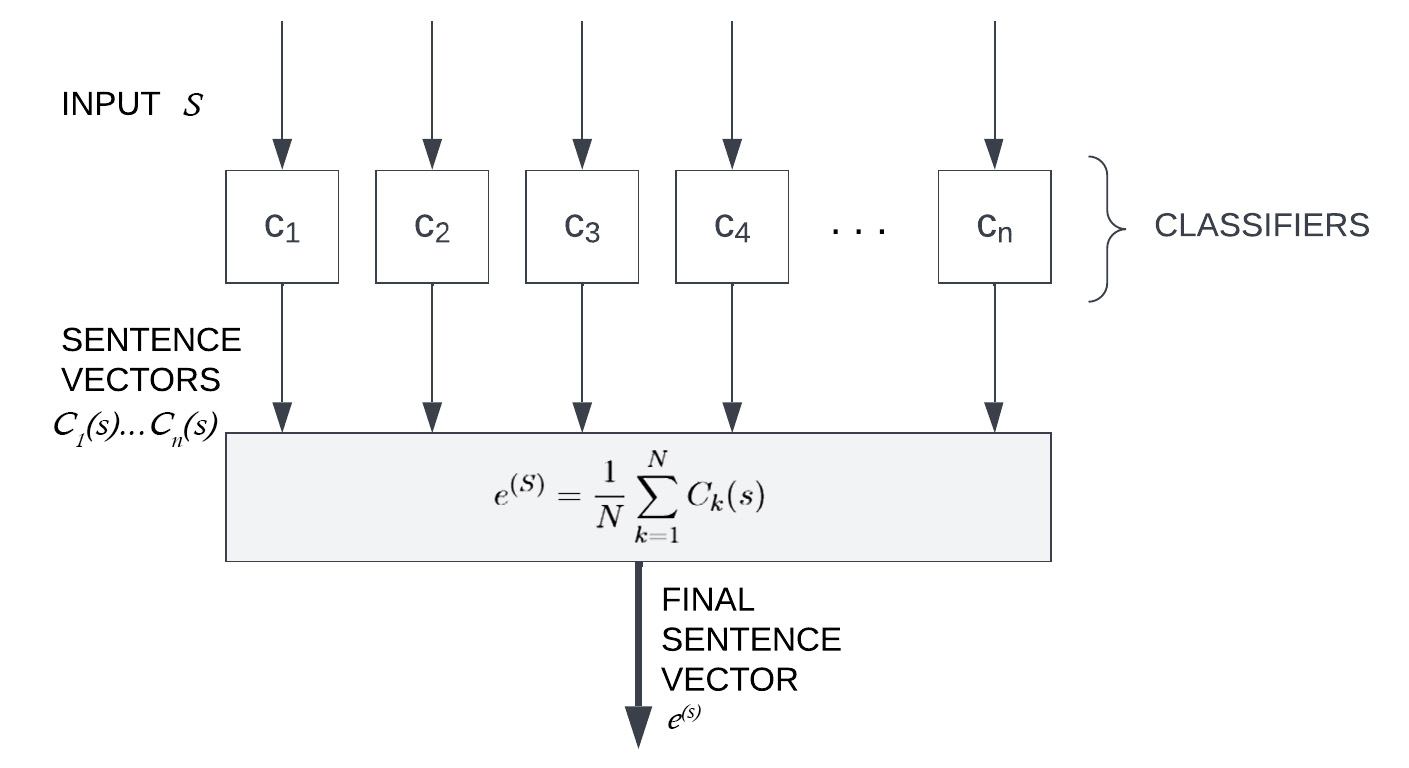}
\caption{Model architecture for SPE: Given input $S$ and $N$ supervised classifiers denoted as $C_1, C_2, \ldots, C_N$. We consider these classifiers to be functions whose outputs are the sentence embeddings extracted from the classifiers when applied to a sentence. The complete embedding of a sentence is obtained by averaging the output of classifiers into a single embedding vector $e^{(S)}$.}
\label{fig:metric_concept}
\end{figure}

Our method is highly flexible and can be adapted to any task depending on the selection of supervised classifiers. Embedding from each classifier is averaged into one, which helps to reduce noise and improve the overall accuracy of a model by smoothing out the influence of any individual data points that may be outliers or contain errors. This process is illustrated in Figure \ref{fig:metric_concept}.

As mentioned above, the use of fastText classifiers allows us to achieve a reduced real-time complexity of SPE compared to USE \cite{USE} or BERTScore \cite{zhangBERTScoreEvaluatingText2020}. SPE only needs to perform $N$ matrix multiplication operations (because we implement $N$ fastText classifiers) with small matrices due to the classifiers hidden layer dimension equal to $H$, which is usually a small number e.g. 10. This is a marginal difference compared to the computations executed by USE \cite{USE}, which performs the operations with $H$-dimensional vectors, where $H$ equals 512 and therefore very large matrices.

The final time complexity to compute the sentence representation using SPE is defined as follows:
\begin{equation}
O=\sum_{i=1}^{N} V*H_{i}
\end{equation}
where $V$ is the length of the sentence and $H_{i}$ is the size of the hidden layer for the i-th classifier.\par

\section{Experiments}
\label{sec:experiments}

To evaluate our proposed sentence encoder in real attack use cases, we performed 
automatic and human-based manual evaluation. As explained in 
\nameref{sec:intro}, the commonly used attack success rate is fundamentally flawed and insufficient to measure the success
of an adversarial attack. \par 
The metric only gives a partial view of the real capability of an attack, obscuring 
the quality of the semantic similarity constraint on the generated sentences. For this reason, we conducted
an extensive survey to collect human evaluations of the quality of the attacks generated, which allows us to evaluate the semantic similarity constraint.

The main goal of our experiments is to study the impact of SPE when used as a semantic similarity 
constraint in adversarial attacks. If this constraint is too strict, only a few high-quality sentences 
would be accepted
as successful attacks, potentially missing many good candidates. Inversely, too loose a constraint 
would produce many low-quality examples.
We focus our experiments on two widely used attacks: TextFooler \cite{jinBERTReallyRobust2020} and its improved 
version TFAdjusted \cite{morris-etal-2020-reevaluating}. Due to the extensive human evaluation, it would be difficult to experiment with all the other adversarial attacks. However, results from these adversarial attacks should be applicable on other adversarial attacks due to their significant similarities.

To understand the effect of SPE in existing 
attacks, we use it as a constraint, together with two other state-of-the-art semantic similarity metrics, 
the Universal Sentence Encoder (USE) \cite{USE} and BERTScore \cite{zhangBERTScoreEvaluatingText2020},
totaling  six attack/sentence encoder pairs.

\subsection{Automatic Evaluation}
\label{sec:automatic_evaluation}
We automatically evaluate adversarial attacks using three metrics:
\begin{description}
    \item[Attack success rate (ASR).] The percentage of successful adversarial 
    examples found by an attack. Given a sentence classifier, a successful adversarial example means that the attack generated a sentence 
    similar to the original that is assigned a different label by this classifier. A higher success rate means that more
    generated sentences are assigned a different label.
  \item[Time.] Average time needed to create an adversarial example. Attacks may rely on various search 
  techniques to generate candidate sentences, leading to varied generation times. 
  \item[Modification rate.] The percentage of modified words. To be as imperceptible as possible, attacks 
  should use as few edits as possible.
\end{description}
  
\subsection{Human evaluation}

\label{sec:ablation_study}
Using automatic evaluation, such as the attack success rate, adversarial attacks are considered successful 
if they simultaneously pass a semantic similarity threshold and manage to change the label of a classifier. 
Measurement of semantic similarity between sentences is still an open research problem with no commonly 
accepted solution. In some cases, even humans disagree on whether two sentences are similar or not.
For this reason, we include an extra evaluation step in our experiments to obtain more robust results.
Sentence pairs were presented to 5 annotators who assigned them an integer score between 1 and 4, where the score 1 means strongly disagree, 2 disagree, 3 agree, and 4 strongly agree.
The higher the score, the more similar the meaning of a pair of sentences will be according to the annotator.
Each annotator was given the sentences to label in several online forms. We averaged the scores of all the 
annotators into a single value for each sentence. Sentences were assigned a binary label; those with an average 
score greater than or equal to 2.5 are considered 
similar, while the others are labeled as not similar. Because the original score system is not binary, the threshold was chosen exactly in the middle of these values. Experimentally, we have determined that the exclusion of the equality to 2.5 does not have any noticeable impact on the results.

Based on this evaluation, we estimate a realistic success rate of the attack --- or an estimated ``real ASR'' (rASR). 
This number represents the number of successful attacks that would actually fool the reader into thinking that 
they have the same meaning as the original sentence, but have a different label assigned by the classifier. 
This score, although costly to estimate, gives a very accurate measure of the quality of an attack. 
A perfect attack would reach a high ASR and a very similar rASR ($\text{ASR} \approx \text{rASR}$), which means that it successfully fooled the 
target classifier repeatedly and that all the corresponding attacked sentences also conserve the same meaning as 
the original ones in the eyes of human annotators. 

\subsection{Datasets}
We evaluated the performance of SPE-based attacks and compared them with existing
state-of-the-art attacks on four text classification datasets that correspond to 
various potential applications of adversarial attacks from NLP, such as data
augmentation or detection of hateful and offensive speech. We did not include sentence similarity benchmarks such as MRPC \cite{mrpc} or other, because they do not contain the types of sentences that occur in adversarial examples. Therefore, they would not be relevant for our context.

\begin{figure}[htbp]
\centering
\vspace{-10pt}
\includegraphics[width=0.6\linewidth]{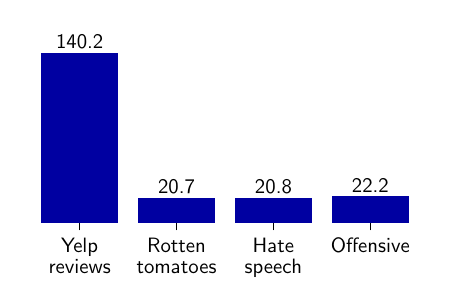}
\caption{Average number of words per sentence in each dataset. The Yelp reviews dataset has much longer
    sentences on average. The two datasets based on Twitter have a built-in character limit (tweet size), 
    and the Rotten tomatoes 
    reviews were pre-processed by the dataset authors to usually contain a single sentence.
\label{fig:wps}}
\vspace{-5pt}
\end{figure}

Figure \ref{fig:wps} 
shows the average number of words in each of the datasets studied.
All source code, links to the datasets, and trained models can be found on \href{https://github.com/DavidHerel/semantics-preserving-encoder}{GitHub}.

\paragraph{Offensive tweets.\label{sec:offensive-tweets}}
This dataset is from the TweetEval
set of seven multiclass tweet
classification tasks \cite{barbieriTweetEvalUnifiedBenchmark2020} . The tasks are labeled irony, hate, offensive, stance, emoji,
emotion,  and sentiment. The offensive task consists of classifying tweets as
offensive or non-offensive. We used a subset of 1000 sentences from the training
split of the dataset for our attacks. This dataset is particularly interesting in 
the context of adversarial attacks on text because it is based on real harmful content 
that can be encountered online. 

The attacked model is a roBERTa-based
\cite{liuRoBERTaRobustlyOptimized2019} model trained on 58M tweets and
fine-tuned for offensive language identification with the TweetEval benchmark
\cite{barbieriTweetEvalUnifiedBenchmark2020}.

\paragraph{Hate speech tweets.} This dataset is from the same set of tasks as the Offensive
Tweets dataset above (Section \ref{sec:offensive-tweets}). The goal of the hate speech task
is to classify tweets as hate speech or non-hate speech. We used a subset of
1000 sentences from the training split of the dataset for our attack. 


The attacked model is a roBERTa-based
\cite{liuRoBERTaRobustlyOptimized2019} model (same as the model used for the offensive tweets dataset)
trained on 58M tweets and fine-tuned for offensive
language identification with the TweetEval benchmark
\cite{barbieriTweetEvalUnifiedBenchmark2020}.

\paragraph{Rotten tomatoes.} The rotten tomatoes dataset is a movie review dataset with
5331 positive and 5331 negative processed sentences from Rotten Tomatoes movie
reviews \cite{pangSeeingStarsExploiting2005}. We used a subset of 1000 sentences
from the training split of the dataset for our attack.

The attacked model is a roBERTa \cite{liuRoBERTaRobustlyOptimized2019} model fine-tuned on the
rotten tomatoes dataset.

\paragraph{Yelp reviews.} The Yelp review dataset is a dataset for binary sentiment classification
\cite{zhangCharacterlevelConvolutionalNetworks2015}. It contains
a set of 560,000 highly polar Yelp reviews for training and 38,000 for testing.
It consists of Yelp reviews extracted from the 2015 Yelp Challenge data.

The attacked model is a roBERTa \cite{liuRoBERTaRobustlyOptimized2019} model fine-tuned on
binary sentiment classification from Yelp polarity.

\subsection{SPE parameters}
Choosing supervised classifiers that are suitable for a wide variety of tasks poses a challenging task. We have trained more than 40 classifiers on different annotated data and performed experiments on a Morris dataset \cite{morris-etal-2020-reevaluating}, which contains 400 annotated adversarial sentence pairs. We have found that 7 selected classifiers achieve reasonably good performance. These classifiers were trained on Natural Language Inference (NLI) datasets like SNLI \cite{snli}, cola, rte and sst2 \cite{glue}, and on more downstream tasks such as emotion classification CARER, Yelp reviews and Stack Overflow questions classification \cite{saravia-etal-2018-carer,zhangCharacterlevelConvolutionalNetworks2015, stack_over_flow}. More details about the selection of classifiers can be seen in Section \ref{classifier_selection}.

Generally, the role of an encoder in a similarity metric is to transform the given text into a corresponding vector in the latent space, which is then used to evaluate the similarity. Specifically, cosine similarity is measured between the vectors of the original text and the perturbed text. If cosine similarity reaches a certain preset threshold $\epsilon$, an adversarial example is considered successful.\par
Therefore, for our metric, the general cosine threshold also had to be defined. Ideally, it should be set so that consistent results are produced for any domain. Our experiments empirically show that setting $\epsilon$ to 0.95 performs consistently well across different datasets. These findings are also similar to thresholds find by Morris et al. \cite{morris-etal-2020-reevaluating}. \par

\subsection{Classifier Selection}
\label{classifier_selection}
Experiments for classifier selection were performed on the Morris dataset \citep{morris-etal-2020-reevaluating}. This dataset contains 400 sentences and their perturbed examples labeled by human annotators, to determine whether it preserves the meaning. Thus making it a valuable dataset for testing our method.
\begin{figure}[h]
\centering
\includegraphics[width=0.88\linewidth]{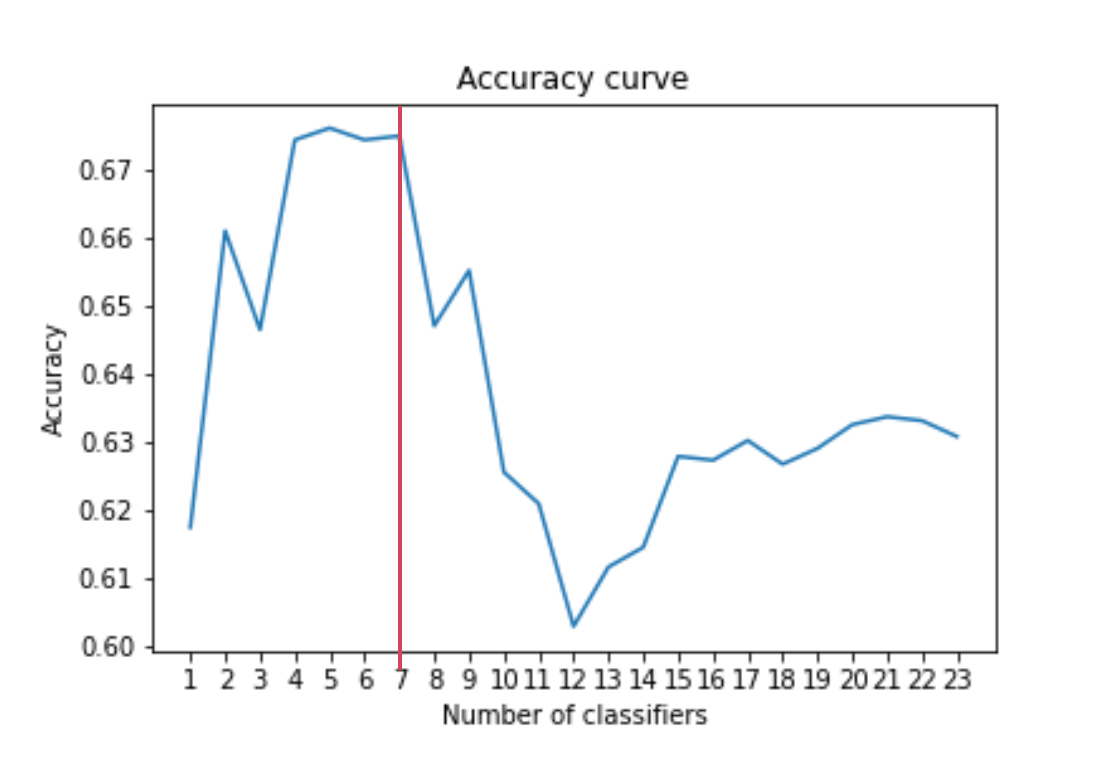}
\caption{Visualization of how addition of other classifiers to the set decreases performance on the Morris dataset \citep{morris-etal-2020-reevaluating}. The reason behind this observation is that some classifiers have learned misleading or irrelevant mapping of words to latent space for our task.}
\label{fig:mrpc_results}
\end{figure}

We have experimentally tried many combinations of sets of classifiers. The best set we have found consists of 7 classifiers as shown in Figure \ref{fig:mrpc_results}. Addition of other classifiers to the set decreased the performance. The reason behind this observation is that some classifiers have learned misleading or irrelevant mapping of words to latent space for our task. \par
An example of such a classifier is one that is trained on a dataset of formal and slang English pairs and the goal is to distinguish between them. Then the classifier will learn to map antonyms and synonyms to a similar latent space and thus will not be able to distinguish between positive and negative sentences, which is what we aim to achieve. This exact case is observed behind the red line in Figure \ref{fig:mrpc_results}.\par
The final set of classifiers was trained on Natural Language Inference (NLI) datasets like SNLI \citep{snli}, cola, rte and sst2 \citep{glue}, and on more downstream tasks such as emotion classification CARER, Yelp reviews and Stack Overflow questions classification \citep{saravia-etal-2018-carer,zhangCharacterlevelConvolutionalNetworks2015, stack_over_flow}. \par

\subsection{Human Participants} 
\label{sec:human_annotators}
Due to the character of the task, human judges had to be advanced in English and understand phrases and shortcuts commonly used online, as well as some cultural references. That is why only native speakers or certified C2-level English speakers were recruited\footnote[4]{According to the international standard - Common European Framework of Reference for Languages (CEFRL).}. The recruitment process took place through a student research group. There were 5 annotators selected in total, who were paid at least a minimum wage, which was agreed upon prior to their involvement.\par
\begin{figure}[h]
\centering
\includegraphics[width=\linewidth]{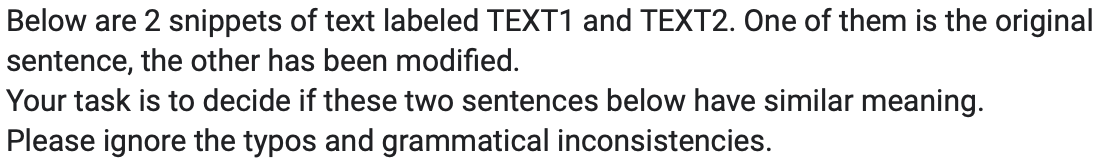}
\caption{Assignment instructions for human judges.}
\label{fig:form_instruction}
\end{figure}
\begin{figure}[h]
\centering
\includegraphics[width=\linewidth]{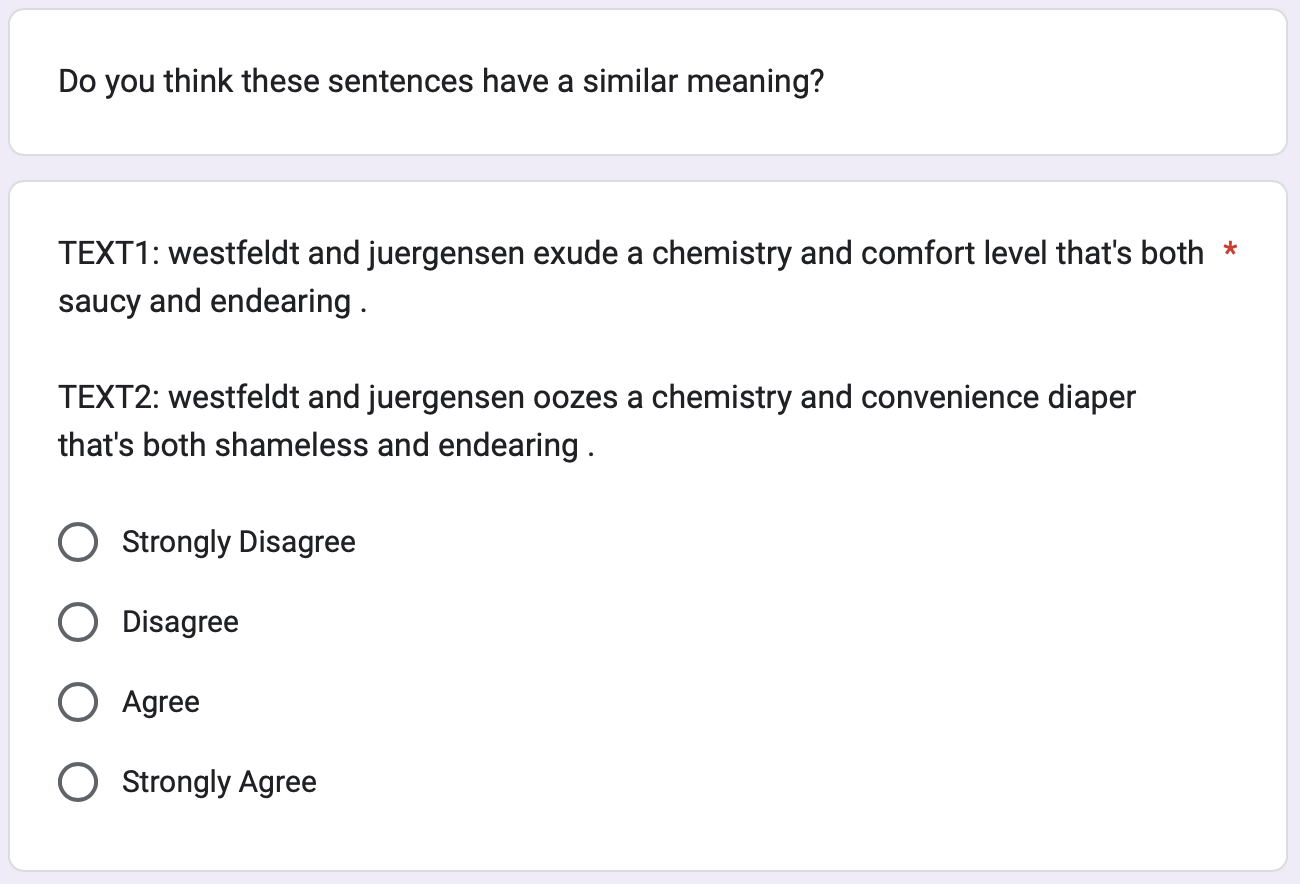}
\caption{A typical page of the form for human judges consists of a sentence pair and selection boxes of available scores.}
\label{fig:form_sentences}
\end{figure}

The human evaluation process took place through multiple online forms (Google Form) - one form for each dataset and attack, which contained an introductory page with the assignment instructions and then one sentence pair on each page (maximum 100 sentence pairs per one form in total) with a selection of scores. For each sentence pair, annotators assigned an integer score between 1 and 4, where the score 1 strongly disagree, 2 disagree, 3 agree, and 4 strongly agree. 
Instructions can be seen in Figure \ref{fig:form_instruction}, and an example of a typical sentence pair is depicted in Figure \ref{fig:form_sentences}.

During the annotation process, annotators were allowed to use the Google Search Engine and an online English dictionary of their choice to clarify any cultural references or shortcuts they may not have been familiar with prior to this task. Other than the exposure of the annotators to offensive online content, which they agreed to undergo voluntarily prior to their participation, there were no risks to the participants. All annotators were thoroughly informed of the task at hand and gave their full consent to participate voluntarily.\par
Annotators had to meet the following criteria: be adults (due to the offensive nature of some of the content for evaluation), and have advanced English (certified as C2 or native speakers). We have asked each annotator to provide proof for each of the above-mentioned criteria. Besides that, no other data about the annotators were collected. All the information collected about human judges is listed in Table \ref{table:annotators_demographics}.

\vspace{-4pt}

\begin{table}
\centering
\begin{tabular}{l|l|l}
 \hline
 Annotator id & Age & English proficiency level\\
 \hline
 1 & 18+ & C2 \\
 2 & 18+ & native speaker \\
 3 & 18+ & native speaker \\
 4 & 18+ & C2 \\
 5 & 18+ & C2 \\
 \hline
\end{tabular}
\caption{Characteristics of human judges.}
\label{table:annotators_demographics}
\end{table}

\newpage
\section{Results}
\label{sec:results}

\begin{table*}[tp]
\centering
\newcommand\mycolsize{0.052\linewidth}
\newcommand\mycolsizebis{0.08\linewidth}
\newcommand\mycolsizetime{0.055\linewidth}
\newcommand\mycolsizemod{0.07\linewidth}
{\scriptsize
\begin{tabular}{p{0.21\linewidth}|p{\mycolsize}|p{\mycolsizebis}|p{\mycolsizetime}|p{\mycolsizemod}p{-20pt}p{\mycolsize}|p{\mycolsizebis}|p{\mycolsizetime}|p{\mycolsizemod}}
 \multicolumn{5}{c}{\normalsize (a) \textbf{Offensive tweets}}&&\multicolumn{4}{c}{\normalsize (b) \textbf{Hate speech tweets}}\\
\cmidrule[\heavyrulewidth]{1-5} \cmidrule[\heavyrulewidth]{7-10}
 \small Attack & \small ASR $\uparrow$ & \small rASR $\uparrow$ & \small Time $\downarrow$ & \small Mod.rate &  & \small ASR $\uparrow$ & \small rASR $\uparrow$ & \small Time $\downarrow$ & \small Mod.rate \\
\cmidrule{1-5} \cmidrule{7-10}
\textbf{\small TextFooler+SPE (ours)} & \small 68.3 & \bfseries \small 30.7 (100) & \small 1.559 & \small 16.8& & \small 65.6 & \bfseries \small 45.9 (100) & \small 1.865 & \small 19.3 \\
\textbf{\small TextFooler+USE} & \small 75.4 & \small 20.4 (100) & \small 1.749 & \small 17.4& & \small 69.1 & \small 27.0 (100) & \small 2.111 & \small 20.9 \\
\textbf{\small TextFooler+BERTScore} & \small 64.2 & \small 18.6 (100) & \small 1.955 & \small 19.5& & \small 65.8 & \small 37.5 (100) & \small 2.274 & \small 20.7 \\
\cmidrule[\lightrulewidth]{1-5} \cmidrule[\lightrulewidth]{7-10}
\textbf{\small TFAdjusted+SPE (ours)} & \small 11.7 & \bfseries \small 9.1 (100) & \small 1.054 & \small 10.6& & \small 10.8 & \bfseries \small 9.8 (97) & \small 1.044 & \small 12.4 \\
\textbf{\small TFAdjusted+USE} & \small 1.0 & \small - 
& \small 0.865 & \small 9.1& & \small 2.6 & \small 2.4 (23) & \small 0.771 & \small 9.7 \\
\textbf{\small TFAdjusted+BERTScore} & \small 5.4 & \small 5.4 (47) & \small 1.167 & \small 10.4& & \small 5.7 & \small 5.7 (51) & \small 1.187 & \small 14.3 \\
 \addlinespace[+\aboverulesep] 
\cmidrule[\heavyrulewidth]{1-5} \cmidrule[\heavyrulewidth]{7-10}
 \multicolumn{10}{c}{}\\
 \multicolumn{10}{c}{}\\
 \multicolumn{5}{c}{\normalsize (c) \textbf{Rotten tomatoes}}&&\multicolumn{4}{c}{\normalsize (d) \textbf{Yelp reviews}}\\
\cmidrule{1-5} \cmidrule{7-10}
 \small Attack & \small ASR $\uparrow$ & \small rASR $\uparrow$ & \small Time $\downarrow$ & \small Mod.rate &  
 & \small ASR $\uparrow$ & \small rASR $\uparrow$ & \small Time $\downarrow$ & \small Mod.rate \\
\cmidrule{1-5} \cmidrule{7-10}
\textbf{\small TextFooler+SPE (ours)} & \small 89.0 & \bfseries \small 61.4 (100) & \small 0.746 & \small 12.8& & \small 87.4 & \bfseries \small 36.7 (100) & \small 14.792 & \small 10.2 \\
\textbf{\small TextFooler+USE} & \small 96.4 & \small 41.5 (100) & \small 0.824 & \small 13.2& & \small 90.5 & \small 7.2 (100) & \small 13.874 & \small 10.6 \\
\textbf{\small TextFooler+BERTScore} & \small 92.7 & \small 38.9 (100) & \small 0.986 & \small 13.9& & \small 89.8 & \small 7.2 (100) & \small 16.176 & \small 11.5 \\
\cmidrule[\lightrulewidth]{1-5} \cmidrule[\lightrulewidth]{7-10}
\textbf{\small TFAdjusted+SPE (ours)} & \small 12.6 & \bfseries \small 11.8 (100) & \small 0.774 & \small 12.4& & \small 8.2 & \bfseries \small 4.5 (81) & \small 36.921 & \small 10.6 \\
\textbf{\small TFAdjusted+USE} & \small 0.3 & \small - 
& \small 0.513 & \small 9.6& & \small 0.3 & 
\small - & \small 22.325 & \small 1.2 \\
\textbf{\small TFAdjusted+BERTScore} & \small 5.7 & \small 5.1 (57) & \small 0.993 & \small 14.2& & \small 2.3 & \small 1.5 (23) & \small 32.946 & \small 5.7 \\
 \addlinespace[+\aboverulesep] 
\cmidrule[\heavyrulewidth]{1-5} \cmidrule[\heavyrulewidth]{7-10}
\end{tabular}
}
\caption{ASR is the attack success rate (in \%), rASR is the estimated real attack success rate obtained 
from the human survey (in \%), the time per sentence is in seconds and the modification rate is the average fraction 
of words changed per sentence (in \%).
An upward arrow ($\uparrow$) indicates 
that higher is better.  All numbers but the estimated rASR were computed on 1000 instances. For the estimated rASR,
we report the number of annotated examples in parentheses. 
For some of the results with TFAdjusted + USE/BERTScore the ASR is so low that few sentences could 
be submitted for annotation, and the rASR estimates are unconclusive.
We omitted the rASR values for configurations with less than 10 successful attacks out of 1000 attempts.
}
\label{table:results_attack_human}
\vspace{5pt}
\end{table*}

The results of our experiments with are shown in Table~\ref{table:results_attack_human}. 
We observe that SPE surpasses other alternatives by a large margin in all configurations, with an estimated real attack success 
rate (rASR) more than 19\% higher, on average, than USE or BERTScore with TextFooler and more than 5 percentage points, on average, higher with TFAdjusted.

As expected, TextFooler-based attacks have a high ASR, reaching 90\% and more while attacking the rotten tomatoes dataset for example. 
But they also have a steep drop from ASR to rASR; for example, TextFooler+USE has its ASR decrease from 96\% to a rASR of 41\%, indicating that many of the supposedly 
successful attacks actually do not have the same meaning according to human annotators. This is due to the weakness of the 
semantic similarity constraints, as they do not filter out low-quality examples. This can be an issue in practical applications, 
since the attacked sentence could easily be detected by humans. For all four datasets, this drop is minimal when using SPE as a constraint.

TFAdjusted receives lower ASR scores than TextFooler (no ASR greater than 12\%), but the drop between ASR and rASR is 
low compared to TextFooler, with a maximum drop of 2.4\% for TFAdjusted+SPE on the offensive tweets 
dataset. It shows that the sentences generated by the TFAdjusted attack are of 
high quality, which is confirmed by the annotators. This is in agreement with the 
observations of Morris et al. \cite{morris-etal-2020-reevaluating} who proposed TFAdjusted to improve the output quality of the TextFooler attack. 
The quality increase is obtained at the expense of the ASR, which can be considerably lower. For example, TFAdjusted with USE obtains ASR scores 
lower than 3\% on all datasets, making it barely usable as an attack, since it generates less than 25 usable sentences out of a 1000 attacked sentences. 
The ASR of SPE is higher than the other constraints, which can be interpreted as SPE being less strict. Yet, our encoder still has 
remarkably low drops from ASR to rASR, and surpasses both other
solutions in all examples in terms of rASR. This indicates a better ability to generate successful attacks that maintain the original meaning of sentences.\par

\begin{table}
    \centering
    \small
    \begin{tabular*}{\linewidth}{m{1.6cm}|m{1.6cm}|m{1.5cm}|m{1.6cm}}
    \toprule
     \bfseries Original sentence & \bfseries Perturbed sentence & \bfseries Cosine similarity (USE) & \bfseries Cosine similarity (SPE, ours)\\
     \midrule
       \textit{This movie is so good}  & \textit{This movie is so \textbf{bad}}  & 0.90 & \bfseries 0.63\\
        \midrule
       \textit{This movie is so good}   & \textit{This movie is so \textbf{tasty}} & 0.89 & \bfseries 0.95\\
       \bottomrule
       
    \end{tabular*}
    \caption{Visualization of a cosine similarity produced by USE \cite{USE} and our approach, SPE. First and second columns represent Original and Perturbed sentence. The third column shows cosine similarity produced by USE and the fourth column shows cosine similarity produced by SPE. Sentences in the first row should have low cosine similarity due to their opposite meaning, and sentences in the second row high cosine similarity due to the similar meaning.}

\label{table:use_spe}
\end{table}

Adversarial attacks using SPE are also much faster than USE or BERTScore, up to 40\% on TextFooler as can be seen in Table \ref{table:results_attack_human}. For TFAdjusted, the generation of adversarial examples may seem to take less time with BERTScore and USE at first glance, however, this is because the generation process fails to produce nearly any successful examples, which causes it to appear faster. These facts are also supported by the observed ASR that was close to 0 for USE and BERTScore. On the contrary, SPE achieves much higher ASR on TFAdjusted, therefore many more successful attacks are generated than with USE or BERTScore. Also, the time complexity of SPE still compares to other approaches while producing valid examples.


\newpage
\section{Conclusion}
\label{sec:conclusion}

We propose a new sentence-encoder technique, SPE, for textual adversarial attacks. Our method outperforms existing sentence encoders used in adversarial attacks by achieving 1.2$\times$ $\sim$ 5.1$\times$ better real attack success rate. Due to the usage of fastText models \cite{fasttext}, SPE is also up to 24\% faster than the existing techniques on TextFooler \cite{jinBERTReallyRobust2020}. 

We have also shown that up to 70\% of the sentences generated by existing state-of-the-art adversarial attacks should be discarded, because they do not preserve their original meaning. With the usage of SPE, the number of incorrect adversarial examples drops remarkably.

Due to the immense growth of online content, the need for defense against adversarial attacks is higher than ever. We demonstrate the core issue of textual adversarial attacks to the community to stay ahead of attackers. Our method, SPE, also addresses this core weakness and helps generate better adversarial examples, which could be used to strengthen existing systems by adversarial training.

Our code is released as a plugin that can be used in any existing adversarial attack to improve its quality and speed up its execution.

\section*{Ethics Statement}
With our proposed technique, SPE, we demonstrate the core issue of textual adversarial attacks to the community to stay ahead of attackers. Our method, SPE, also addresses this core weakness and helps generate better adversarial examples, which could be used to strengthen existing systems through adversarial training.

Our method also improves the efficiency of adversarial attack generation, and thus is environmentally friendly. We achieve up to 24\% speed-up compared to the SOTA encoders on TextFooler \citep{jinBERTReallyRobust2020}. We also release the source code to make the application of our encoder efficient and as easy as possible. 

\ack We would like to thank Daniela Hradilova for the useful discussions and suggestions. Our work is part of the RICAIP project that has received funding from the European Union’s Horizon 2020 research and innovation programme under grant agreement No. 857306.

\bibliography{spe_paper_david, spe_paper_hugo}

\newpage

\end{document}